\documentclass[letterpaper, 10 pt, conference]{ieeeconf}  % Comment this line out if you need a4paper
\IEEEoverridecommandlockouts                              % This command is only needed if 
\usepackage{graphics} % for pdf, bitmapped graphics files
\usepackage{epsfig} % for postscript graphics files
\usepackage{times} % assumes new font selection scheme installed
\usepackage{amsmath} % assumes amsmath package installed
\usepackage{amssymb}  % assumes amsmath package installed
\usepackage{mathtools}
\DeclareMathOperator*{\argmax}{arg\,max}

\newcommand{\triangulation}{T}
\newcommand{\pixel}{x}
\newcommand{\pixels}{\mathbf{\pixel}}
\newcommand{\referencePixel}{\pixel^{ref}}

\newcommand{\segmentation}{S}
\newcommand{\segmentationPrior}{S_P}
\newcommand{\disparity}{d}

\newcommand{\allPriors}{\triangulation, \segmentationPrior, \referencePixel}

\newcommand{\disparityDistribution}{p(\disparity | \triangulation, \referencePixel)}
\newcommand{\pixelLikelihood}{p(\pixels|d,\referencePixel,\segmentation)}

\usepackage{cleveref}
\usepackage{xcolor}
\usepackage{multirow}
\usepackage{array}
\usepackage{caption}
\usepackage{subcaption}
\usepackage{changes}
\usepackage{todonotes}

\title{\LARGE \bf
A Light Field Front-end for Robust SLAM in Dynamic Environments
}

\author{Pushyami Kaveti$^{1}$ and Hanumant Singh$^{2}$% <-this % stops a space
\thanks{$^{1}$Pushyami Kaveti is a student at Khoury College of Computer Sciences, Northeastern University, Boston
        {\tt\small kaveti.p@husky.neu.edu}}%
\thanks{$^{2}$Hanumant Singh is with the Department of Electrical Engineering,Northeastern University
        {\tt\small ha.singh@northeastern.edu}}%
}

\begin{document}
\maketitle
\thispagestyle{empty}
\pagestyle{empty}

%%%%%%%%%%%%%%%%%%%%%%%%%%%%%%%%%%%%%%%%%%%%%%%%%%%%%%%%%%%%%%%%%%%%%%%%%%%%%%%%

\begin{abstract}
There is a general expectation that robots should operate in urban environments often consisting of potentially dynamic entities including people, furniture and automobiles. Dynamic objects pose challenges to visual SLAM algorithms by introducing errors into the front-end. This paper presents a Light Field SLAM front-end which is robust to dynamic environments. A Light Field captures a bundle of light rays emerging from a single point in space, allowing us to see through dynamic objects occluding the static background via Synthetic Aperture Imaging(SAI). We detect apriori dynamic objects using semantic segmentation and perform semantic guided SAI on the Light Field acquired from a linear camera array. We simultaneously estimate both the depth map and the refocused image of the static background in a single step eliminating the need for static scene initialization. The GPU implementation of the algorithm facilitates running at close to real time speeds of 4 fps. We demonstrate that our method results in improved robustness and accuracy of pose estimation in dynamic environments by comparing it with state of the art SLAM algorithms.
%Our method results in improved robustness and accuracy of pose estimation in situations where the dynamic objects occupy significant fraction of the scene. We demonstrate the effectiveness of our method on real-world data by comparing its performance with the state of the art visual SLAM algorithms.
\end{abstract}
\section{Introduction}
\begin{figure*}[ht!]
\centering
\includegraphics[clip,trim={0cm 2cm 0cm 1cm}, width=0.9\linewidth]{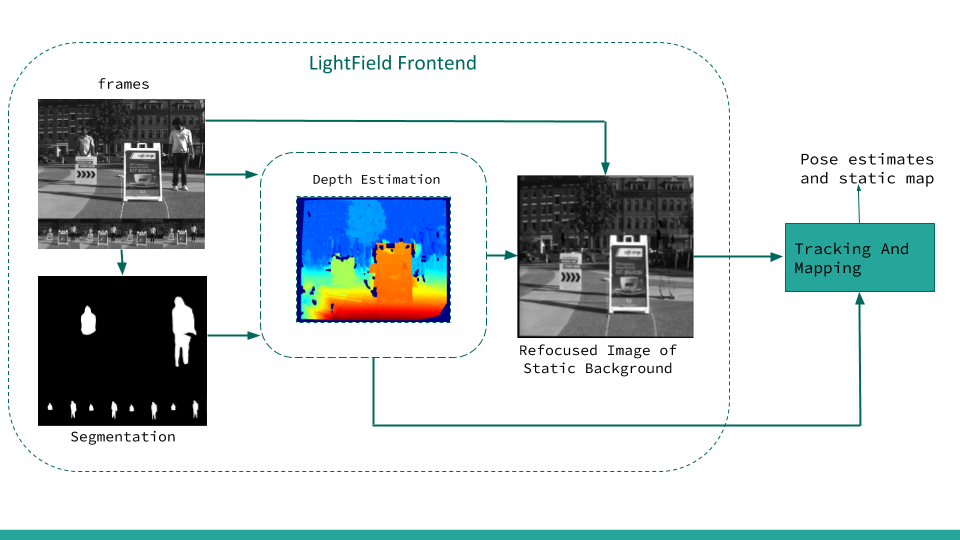}
\caption{Block diagram of the Light Field Front-end. Frames from the linear camera array and the segmentation masks are fed into the depth estimation module. The optimized depth is used to compute the refocused image.}
\label{fig:blk}
\end{figure*}
SLAM enables robot navigation in unknown environments by accurately estimating the robot’s pose and building the map of the world at the same time. It is essential for collision-free navigation in various indoor and outdoor robotic applications. Most of the research in SLAM assumes a static world, even though the robots are expected to function in predominantly dynamic environments. The static world assumption holds true for some applications during small scale and short-term runs, but limits the robots from operating in the real-world. The presence of dynamic entities like people, automobiles and bicycles in the real world causes errors in various stages of SLAM if these objects are not detected and dealt with. Feature matches associated with dynamic objects result in initialization failures, errors in pose estimation and incorrect maps of the world. If the dynamic features are included in the map they affect loop closure and re-localization preventing re-usability of maps for long-term applications. All these problems have motivated researchers to develop algorithms that target dynamic environments~\cite{Cadena2016}\cite{Saputra2018b}.

Dynamic environments are usually handled by detecting dynamic content in a variety of ways \cite{Mur-Artal2017a}\cite{Tan2013b}\cite{Klein2007}\cite{Alcantarilla2012a}\cite{Bescos2018a}\cite{Yu2018b} and explicitly omitting the features associated with the detected dynamic objects while performing SLAM. Discarding information this way may result in errors if the dynamic content occupies a significant portion of the image or dominates the scene in terms of feature space. Therefore, extracting static features is imperative to estimating the pose of the robot accurately.

In this paper, we introduce a Light Field front-end built on top of ORBSLAM2\cite{Mur-Artal2017a} that is robust to dynamic environments. The front-end aims at accurate pose estimation and builds the static map of the world for long term use. While Light Fields have been researched extensively in the computer graphics community, their applicability to mobile robotics is far more recent\cite{Dong2013}\cite{Bajpayee2019}\cite{Zeller2018}. A Light Field captures spatial and angular radiance in the form of a bundle of rays coming from a point in space as opposed to a single ray in a monocular camera. This helps us to extract the rays emerging from partially occluded portions of the scene via Synthetic Aperture Imaging (SAI).

We apply SAI to the Light Field collected from a camera array to synthesize the refocused image of the static background occluded by dynamic objects which is then passed on to extract features for tracking. We rely on deep learning based semantic segmentation for detecting a priori dynamic objects. We use the dynamic object detections to guide the SAI and recover the static background image essentially seeing through the dynamic objects as if they were never present at all. Thus, our method is effective even when the dynamic objects dominate the scene. Detecting a priori dynamic objects supports long term mapping when the objects are temporarily static and eliminates the need for initialization if they are actively moving. Our GPU implementation allows us to run the front-end processing close to real-time. In the next sections we discuss the Light Field front-end in detail and evaluate it on real datasets.
%We apply SAI to the Light Field collected from a camera array to synthesize the refocused image of the static background occluded by the dynamic objects. This refocused image is then passed on to extract features for tracking. We pose the problem as a probabilistic graphical model where we jointly estimate the depth map and the refocused image of the static background given the dynamic objects. We rely on deep learning based semantic segmentation for detecting a priori dynamic objects.Instead of simply discarding the detected dynamic content, we use that information to guide the SAI to compute the background image seeing through the dynamic objects as if they were never present. Thus, our method is effective even when the dynamic objects dominate the scene. Detecting a priori dynamic objects supports long term mapping when the objects are temporarily static and also eliminates the need for initialization if they are actively moving.Our GPU implementation allows to run the front-end processing close to real-time. In the next few sections we discuss the related work, explain the front-end in detail and evaluate our Light Field front-end on real datasets.

\section{Related Work}
%what we cite depends on what we are doing in the paper
% trying to eliminate the initialization phase, dynamic scene SLAM solutions
% at each time step not just detecting the dynamic objects and discarding them, but extract static features as they are crucial for stable localization.

%In this section we discuss some of the recent work in the area of multi-view and light field based reconstruction in the context of robotic applications that closely relates to our method.
\textbf{\textit{SLAM in Dynamic Environments}}
A large body of recent research addresses dynamic scenes by using optical flow \cite{Alcantarilla2012a}, RANSAC\cite{Mur-Artal2017a}, or residual cost functions \cite{Klein2007}\cite{Scona2018a} to detect and filter dynamic outliers. Outlier rejection strategies fail when the dynamic content constitutes a significant fraction of the scene. Other algorithms enforce temporal and structural consistency to detect dynamic content and omit such content from tracking and mapping. In \cite{Kim2016a}, the background is reconstructed by accumulating warped depth between consecutive RGB-D images. \cite{Tan2013b} track changes by projecting sparse map points into the current frame. StaticFusion~\cite{Scona2018a} detects moving objects and fuses temporally consistent data via model alignment. Temporal methods require initialization of static map in dynamic environments. We use CNN based semantic segmentation\cite{he2017mask}\cite{bodypix}, which
has emerged as an efficient instantaneous approach to detect apriori dynamic objects in many recent works\cite{Yu2018b}\cite{Bescos2018a}\cite{Barsan2018a}\cite{palazzolo2019refusion}\cite{zhang2020vdoslam}. Unlike other methods, we not only track features belonging to the statically segmented parts of the image but also extract and track the static features occluded by dynamic objects from the refocused image of the static background. Thus, we demonstrate increased accuracy and robustness even when the dynamic content starts to dominate the scene. \\
%Detecting static features still remains an important task for localizing the robot. In this paper we focus on reconstructing the static background by not simply discarding the dynamic portions but seeing through them and synthesizing a refocused image using Light Fields. We combine the semantic information from segmentation as prior knowledge about the dynamic object and also use multi-view geometry to reconstruct the static scene.\\
\textbf{\textit{Light Fields in Robotics}}
Light Fields are a popular topic in computer vision and graphics for refocusing and rendering\cite{Levoy1996}\cite{Isaksen2000}\cite{Lumsdaine2010}\cite{Davis2012}, with applications that include  super-resolution\cite{Bishop2012} and depth estimation\cite{Vaish2006b}\cite{Wanner2012}\cite{Wang2015}\cite{Williem2018b}\cite{Sheng2018a}. Application of Light Fields in robotics is not as developed but has been discussed in\cite{Dong2013}\cite{Dansereau2011}\cite{Bajpayee2019}. Most of the robotics related work aims to solve the visual odometry problem~\cite{Dansereau2011}, and to compensate for challenging light and weather conditions\cite{Skinner2016}\cite{Bajpayee2019}\cite{Zeller2018}\cite{Eisele2019}. None of this work addresses the problem of dynamic environments. In contrast, our efforts in semantic guided refocusing and reconstruction specifically deal with dynamic objects in a SLAM framework. Refocusing\cite{Isaksen2000} is performed by blending rays emerging from a focal surface and passing through a synthetic aperture without taking into account the semantics of the rays. The graphics community's Light Field depth reconstruction techniques typically are aimed at increasing the accuracy of reconstruction while trading off speed of computation. Most of the Light Field research is targeted towards micro lenslet cameras\cite{Ng2005} which suffers from small baseline separation, which are far from ideal in terms of seeing through the dynamic objects, where wide-baseline arrays are more suitable. %Hence, we use a linear array of five cameras similar to~\cite{Bajpayee2019}, that is most suitable for deploying on robots.
%The main goal of the project is to extract good, trackable features for performing slam in dynamic environments without having to rely on an initial static map; Thus making the robot navigate to any unknown location and build the map seamlessly.
\section{Light Field front-end}
At the heart of the Light Field front-end is our method to compute an all-in-focus synthetic aperture image of the static background from the Light Field captured with a linear camera array.
%given the dynamic object detections. The synchronized frames from the calibrated camera array are run through MaskRCNN\cite{he2017mask} for dynamic object segmentation. We show results of our algorithm by detecting people because they constitute the most common class of dynamic objects; but the method is agnostic to the segmentation class.% The segmentation masks and frames  are used to compute the depth map of the static background. The refocused image of the background is then computed on a per-pixel basis using the background depth map and semantic masks. 
The processing pipeline is illustrated in \cref{fig:blk}.
\subsection{Problem Setup}
\begin{figure}[ht]
\smallskip
\centering
\includegraphics[scale=0.28, clip, trim={0 1cm 0 0cm}]{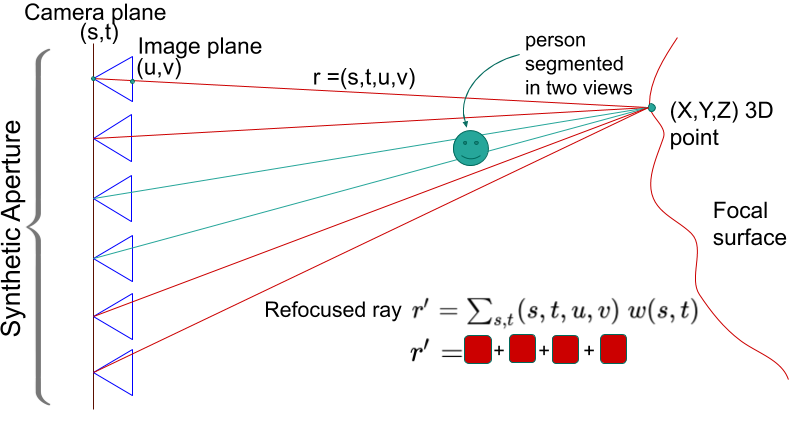}
\caption{Light Field refocusing. Given a 3D point on the focal surface the static rays projected into other cameras are combined to create a single refocused image.}
\label{fig:synaperture}
\end{figure}

A calibrated linear array of K cameras forming images $I_k$ can be seen as a single system with synthetic aperture where each camera contributes to a ray passing through the aperture. Following the two-plane parameterization of Light Field~\cite{Isaksen2000}, the camera positions on the array are the (s,t) coordinates which lie on the entrance plane and the pixels are the (u,v) coordinates on the image plane. This creates a unique 4D ray (s,t,u,v) as shown in \cref{fig:synaperture}. Given an arbitrary focal surface F and a 3D point (X,Y,Z) on it, we can combine all the rays of the synthetic aperture which emit from that point and intersect the cameras into a single refocused ray. 

Suppose we have the depth map of the static background. Let $\disparity^{ref}$ and $\referencePixel$ be the depth and coordinates of a pixel in the reference view, then the rays $\pixel^k$ in the other cameras $k$ can be sampled using:
\begin{equation}
x^k = \pi_k[R_k|t_k]\pi_{ref}^{-1}(x^{ref}, d^{ref}) \label{eq:1}
\end{equation}
where $\pi_k$ is a projective mapping between the a 3D point in space and 2D pixel coordinates on the image plane, and $R_k$ and $t_k$ are the rotation and translation of the $k^{th}$ camera. These rays form our synthetic aperture and are typically combined by applying an average filter\cite{Lumsdaine2010,Isaksen2000} giving equal weight to all the rays.This helps to generate images with varying focus and depth of field \cite{Levoy1996}\cite{Davis2012}. In free-space, all the camera rays correspond to the same point on the focal
surface, but, in case of occlusions, some rays get obstructed. Given a large synthetic aperture, choosing a focal surface behind the foreground object brings the background into focus and causes the object to blur. This was applied in ~\cite{Bajpayee2019} to see through rain and snow by manually selecting a focal plane on the road signs for autonomous navigation. Instead of averaging all the rays, we design a filter which selects only the rays that map to static pixels in the respective views obtained from semantic segmentation. This way the dynamic objects in the foreground are completely eliminated. Denoting whether a pixel $\pixel^k$ belongs to the static or dynamic content with $s^k$, the refocused image $I^*$ can be calculated as follows:
\begin{equation}
I^s = \frac{\sum I(x^k) * (s^k = static)}{\sum (s^k = static)}\label{eq:2}
\end{equation}
As we can observe in~\cref{eq:1}, the depth map of the static background is required to compute the refocused image. %Since manual selection \cite{Bajpayee2019} of the static focal surface is not suitable for real time applications we calculate the depth map of the static background as detailed in next sections.
\subsection{Probabilistic Graphical Model} \label{sc:graphical model}
\begin{figure}[ht]
\centering
\includegraphics[scale=0.48, trim={0cm 0cm 0cm 0cm}]{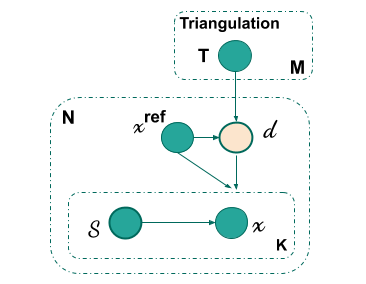}
\caption{The Probabilistic Graphical Model. Triangulation $T$ and segmentation $S$ are pre-computed. $d$ is disparity of a pixel in the reference view, while $x$ is a set of K reprojections into other cameras.}
\label{fig:pgm}
\end{figure}

The main challenge in the depth estimation of the static background is to reconstruct parts of the reference view occluded by dynamic objects. Given that an array of cameras are acting as the source of image data, we need to estimate the depth map $D^{*}$ and refocused image $I^*$ of static background for a reference camera view $X^{ref}$. The key to finding correct depth is to choose the subset of camera views which will exclude the pixels corresponding to dynamic objects while computing the image correspondences. We apply MaskRCNN\cite{he2017mask} on camera images to compute per-pixel semantic labels. The CNN model assigns a probability to pixels being segmented $\{ \segmentationPrior \in \mathbb{R}^k | 0 \leq \segmentationPrior^k \leq 1 \}$, where $\segmentationPrior^k$ is the probability in $k^{th}$ view. We introduce a set of binary random variables $S^k$ assigned to each pixel \textit{i} in camera \textit{k} which take values either 0 or 1 representing dynamic or static pixels based on a threshold on the probability.
The probabilistic graphical model is shown in the \cref{fig:pgm}. The disparity $\disparity$ can be obtained by maximizing the posterior probability $p(d|\pixels,\referencePixel,T,Sp)$ (MAP estimate) as follows:
\begin{align*}
d^* = argmax_d \;\; p(d|\pixels,\allPriors), \\
p(d|\pixels,\allPriors) \propto p(\disparity,\pixels | \allPriors) 
\end{align*}

\begin{figure*}[ht!]
%\smallskip
\centering
\begin{tabular}{ccc}
\subfloat[]{\includegraphics[width = 0.29\linewidth]{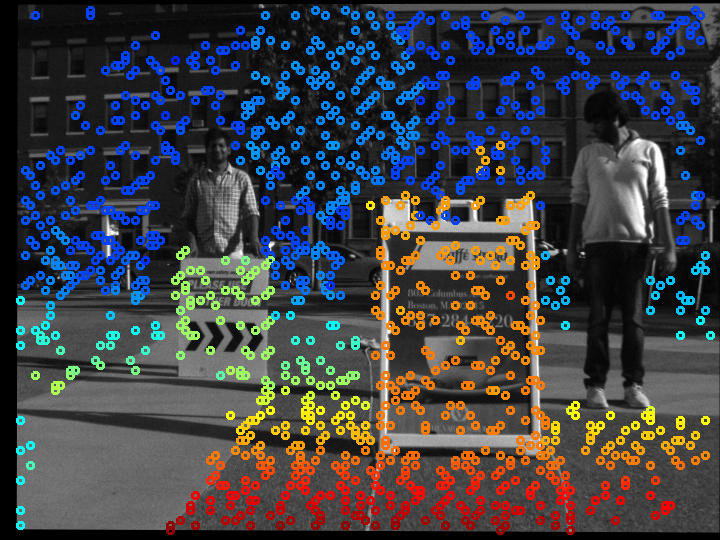}} &
\subfloat[]{\includegraphics[width = 0.29\linewidth, clip, trim={1cm 0cm 1cm 0cm}]{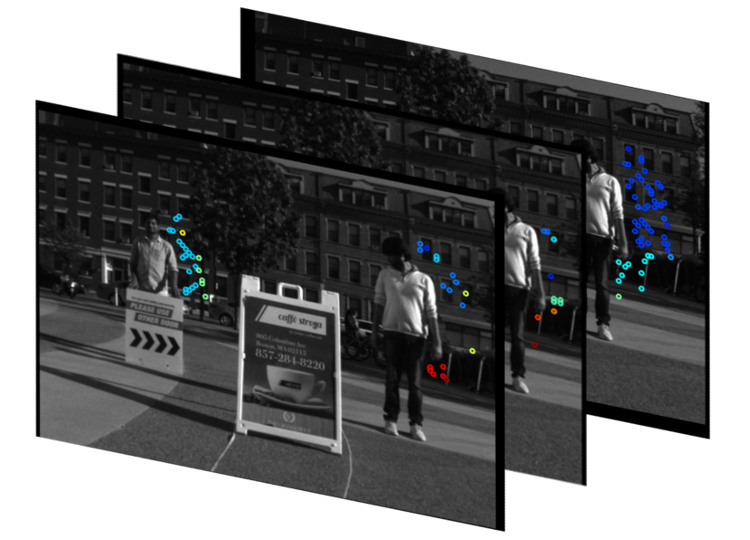}} &
\subfloat[]{\includegraphics[width = 0.29\linewidth, clip, trim={1cm 0cm 1cm 0cm}]{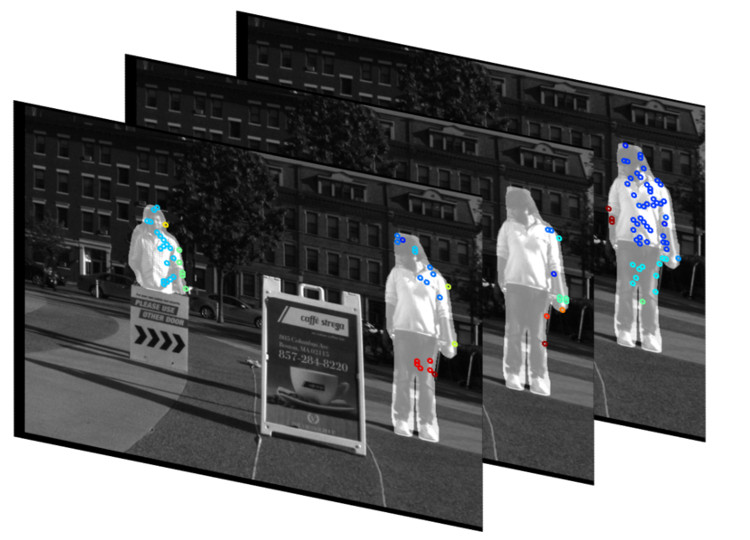}} \\
\subfloat[]{\includegraphics[width = 0.29\linewidth]{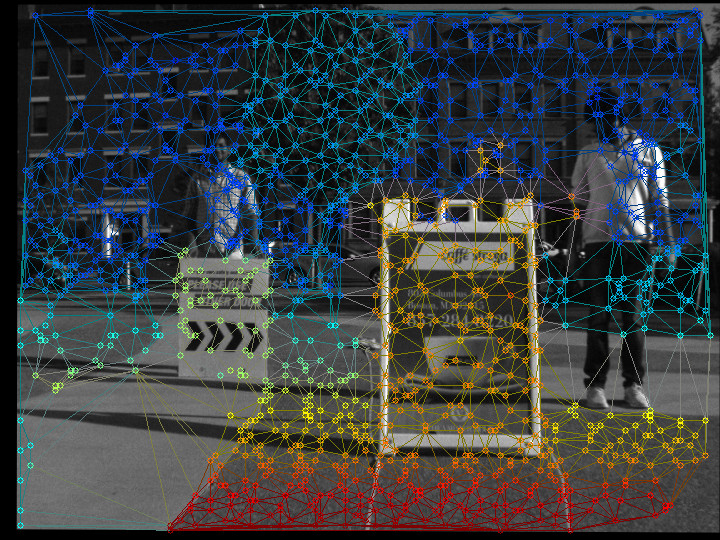}} &
\subfloat[]{\includegraphics[width = 0.29\linewidth]{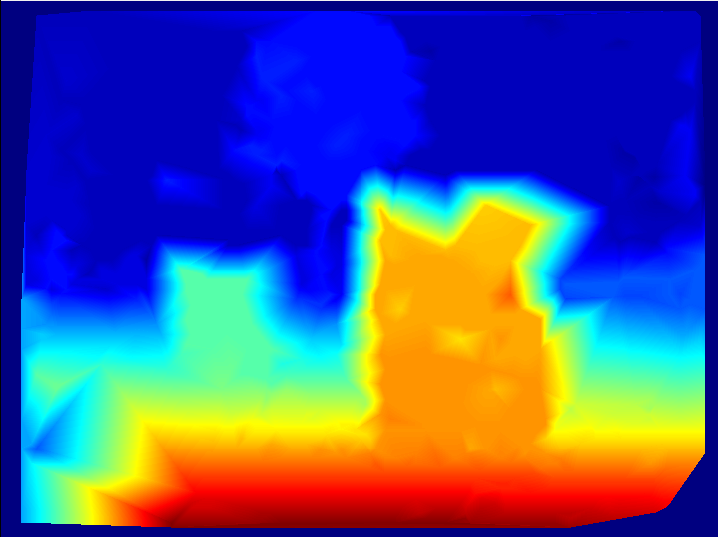}} &
\subfloat[]{\includegraphics[width = 0.29\linewidth]{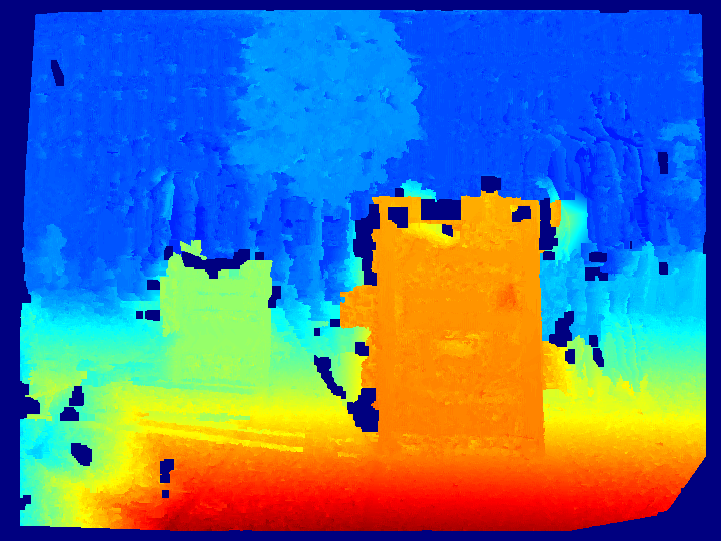}}
\end{tabular}
\caption{Various steps associated with the static background disparity estimation. (a) A sparse set of support points only on the static background in the reference view. (b) Support points from other views. (c) Reprojected support points in the reference frame. The support points are colored based on their disparities (red: close, blue : Far). (d) Final triangulation of the static background. (e) The coarse disparity map formed by the piece-wise planar prior. (f) The final refined disparity map of the static background.}
\label{fig:depth}
\end{figure*}
\subsection{Disparity Estimation}
According to the graphical model, the joint probability factorizes into disparity prior and pixel likelihood. Now, disparity estimation is a task of picking a value that optimizes:
\begin{equation} \label{eq:mstep}
    \argmax_\disparity \; \log\underbrace{\pixelLikelihood}_{\text{pixel likelihood}} + \log\underbrace{\disparityDistribution}_{\text{disparity prior}}
\end{equation}
% complete log likelihood term
% Disparity prior factor
\subsubsection{Disparity Prior}
We use piece-wise planar prior over the disparity space $\disparityDistribution$ by forming triangulation on a sparse set of points similar to ELAS~\cite{geiger2010efficient}. This helps with poorly textured regions and gives a coarse disparity map, thus reducing the search space during optimization.

First a sparse set of points are selected on a regular grid in the reference image and their descriptor vectors are created from the responses of a 5x5 window sobel filter. Since some of these points lie on the dynamic objects we use segmentation mask to filter them. The remaining points are matched along the full range of disparities on the epipolar lines of its neighbouring image. These points together with their matched disparities form the static support points. We now exploit the Light Field data to extract support points of the static background occluded by the dynamic objects in reference view, but are visible in other views. For each image other than the reference view we extract the support points by matching the sobel descriptors of the static points 
against its neighbouring view and choose the points that re-project onto the parts of the reference image dynamic objects. We then filter out duplicates and inconsistent support points based on their disparity values. The final set of support points are used as vertices for the delaunay triangulation. The prior on disparity $\disparityDistribution$ is modeled as combination of a uniform distribution and a sampled Gaussian centered about the interpolated disparity from the triangulation, for support points in the neighbourhood\cite{geiger2010efficient}. All the above steps are shown in \cref{fig:depth} (a-e). 
\subsubsection{Pixel Likelihood}
The triangulation prior provides a coarse map of the interpolated disparities and a set of candidate disparities. which we have  used to generate  accurate estimates based on the pixel likelihood. Given the coordinate $\referencePixel$ in the reference frame and a candidate disparity $d$, the corresponding coordinates of the source images $x^k$ can be determined using a warping function $\mathcal{W}_k(x^{ref}, d)$. This warping function represents a homography which can be computed from the reference and $k^{th}$ camera matrices~\cite{szeliski1999stereo}. This allows us to use generalized disparity space suitable for multi-view camera configuration. %as opposed to a classic multi-baseline setup where cameras are perfectly placed in a plane perpendicular to their optical axes.
We work in the disparity space as opposed to depth space because it facilitates discrete optimization that results in faster computation. The pixel likelihood relies on the fact that for correct disparity value there is a high probability that warped static pixels $x^k$ in the source images will have photometric consistency. So, we design our likelihood function as a Laplace distribution restricted to static pixels such that the variance between them is minimum.
\begin{equation*}
p(x^1...x^K \mid d,x^{ref},S) \\
\propto \begin{cases} exp\big( -\beta\; Var(f(x^1),...,f(x^K))\big)\\
\;\;\;\;\; for \;\;  x^k = \mathcal{W}_k(x^{ref}, d)\; \\
0 \;\;\;\;\; otherwise
\end{cases}
\end{equation*}\\
Where the variance is computed on the feature descriptors $f(x^K)$ of the pixels that are classified as static:
\begin{equation*}
Var(...) = \frac{\sum_{k=1}^K (S^k=1)[f(x^k)-\hat{f}]^2}{\sum(S^k=1)}
\end{equation*}
with $\hat{f} = \frac{\sum_k (S^k=1)f(x^k)}{\sum_k(S^k=1)}$ as the mean of the descriptors. We use the descriptors created from 3x3 sobel filter responses similar to ELAS\cite{geiger2010efficient}. The dynamic pixels don’t contribute to the likelihood as they can have arbitrary intensities. Plugging in the disparity prior and pixel likelihood into \cref{eq:mstep} and taking negative logarithm gives the energy function to be minimized
% M-step energy function
%\begin{align*}
%    E(\disparity) &= \beta\; Var(f(x^1),...,f(x^k)) \nonumber \\ 
%    &- log\bigg[\gamma +exp\Big( -\frac{\big[d-\mu(T,x^{ref})\big]^2}{2\sigma^2}\Big) \bigg]
%\end{align*}
\begin{figure*}[ht]
\medskip
\centering
\includegraphics[clip, trim={0cm 2cm 0cm 0cm},width=0.31\linewidth]{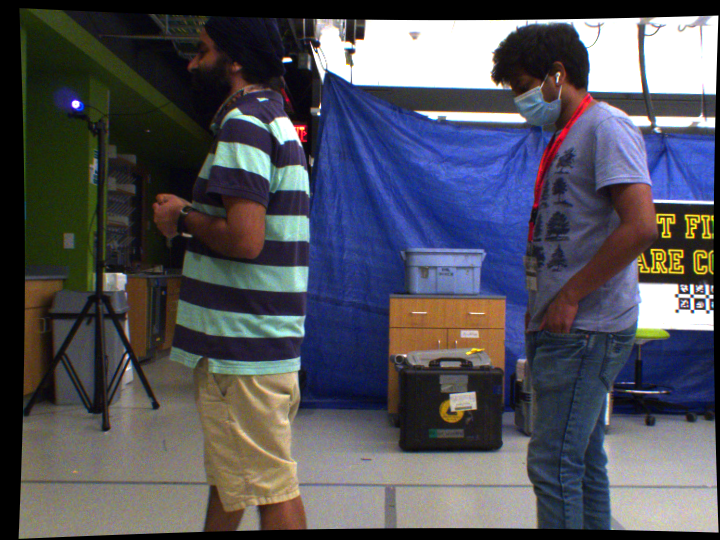}
\includegraphics[clip, trim={0cm 2cm 0cm 0cm},width=0.31\linewidth]{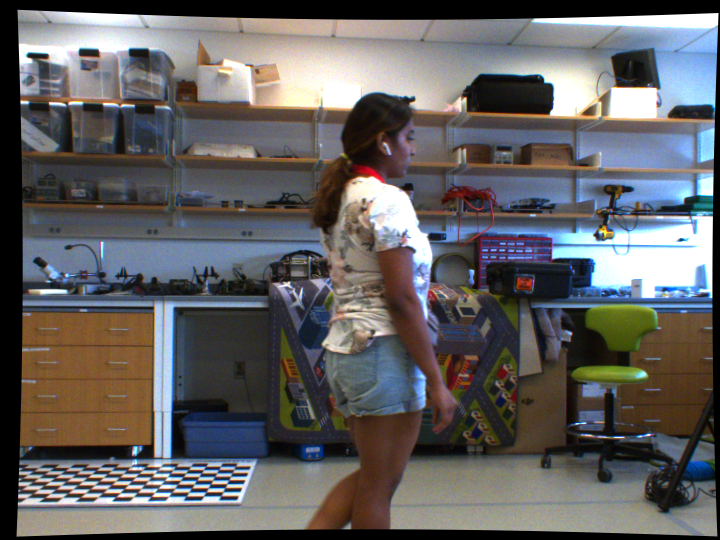}
\includegraphics[clip, trim={0cm 2cm 0cm 0cm},width=0.31\linewidth]{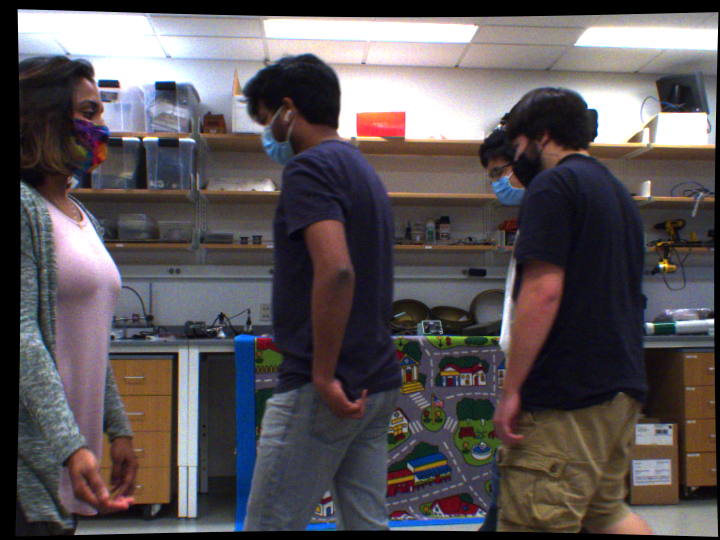}
\includegraphics[clip, trim={0cm 2cm 0cm 0cm},width=0.31\linewidth]{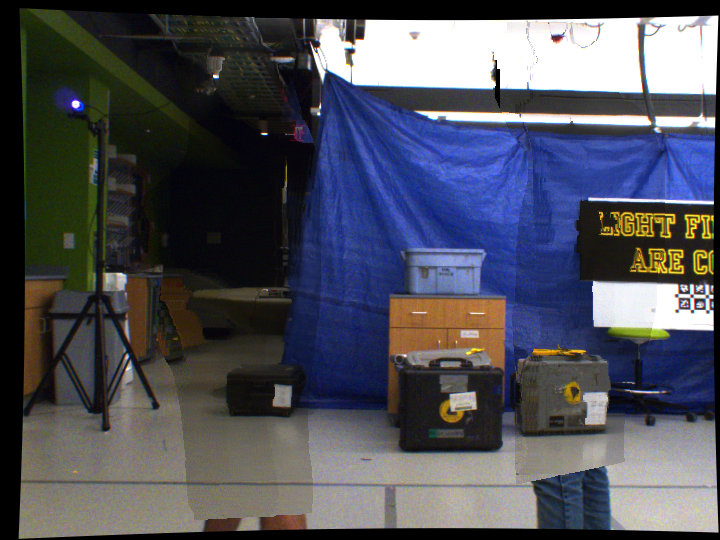}
\includegraphics[clip, trim={0cm 2cm 0cm 0cm},width=0.31\linewidth]{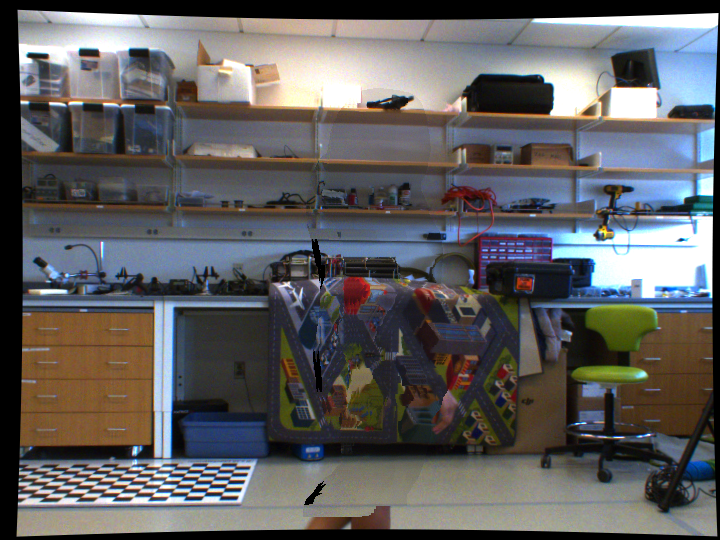}
\includegraphics[clip, trim={0cm 2cm 0cm 0cm},width=0.31\linewidth]{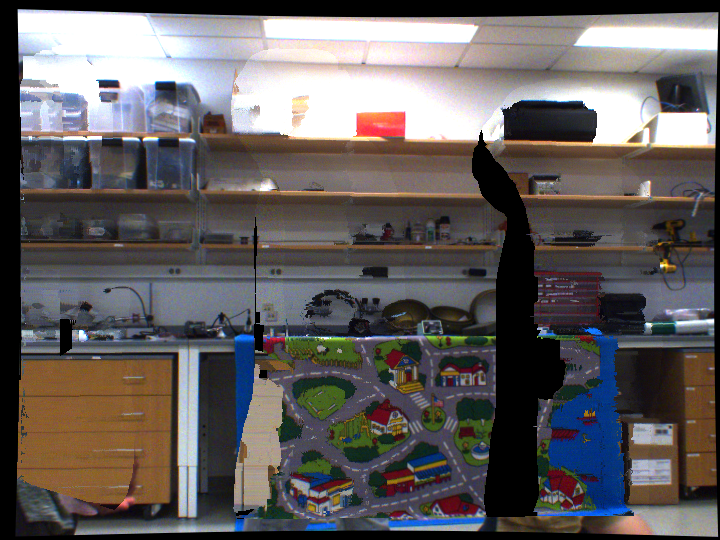}
\caption{\textit{Top row:} Original images in the reference view. \textit{Bottom row:} Refocused image of the static background. First column shows the result with some low textured regions. The second column shows the same in the presence of a lot of detail. Third column shows the case when dynamic objects dominate the scene.}
\label{fig:refocus_res}
\end{figure*}
\subsection{Refocused Image Synthesis}
Once the optimization converges we perform gap interpolation and median filter on the estimated disparity image to get rid of any speckle noise obtain a smooth map. The filtered disparity and segmentation masks are used to compute the refocused image of the static background using \cref{eq:2}.

\section{Experimental Analysis}
In this section, we evaluate the Light Field front-end on real-world dynamic sequences. %by comparing it with the state-of-the-art SLAM solutions for various sensing modalities - Mono, stereo and RGB-D
\\\textbf{\textit{Dataset Acquisition and Setup}}
%\begin{figure}[tb]
%\centering
%\includegraphics[width=0.7\linewidth, clip, trim={0 4cm 0 2cm}]{figures/husky_BG.jpg}
%\caption{The data collection system with the custom-built linear camera array mounted on husky.The cameras are hardware synced with a master-slave architecture.}
%\label{fig:husky}
%\end{figure}
Light Field acquisition is typically done via a large array of cameras~\cite{Levoy1996} or a micro lenslet camera~\cite{Ng2005}. The former is impractical to mount on most mobile robots where as the latter suffers from limited parallalax due to small baseline separation. Based on these constraints we use a custom built 5-camera linear array and mount it on a Clearpath Robotics Husky UGV running ROS for collecting Light Field data. The array uses 1.6MP Pointgrey BlackflyS global shutter cameras which are hardware synced for synchronized image capture and calibrated using the Kalibr\cite{furgale2013unified} package. All our experiments were conducted on an Alienware gaming laptop with a 8GB GEForce GTX 1080 GPU and Intel core i7 processor. We collected multiple indoor datasets with people moving in the scene to depict dynamic environments. The ground truth trajectories were obtained using optitrack prime x13 cameras \cite{optitrack} which are accurate to a millimeter.

\textbf{\textit{Qualitative results}} We show the results of our refocused image synthesis on a few Light Field snapshots from our datasets in \cref{fig:refocus_res}. The top row shows the original images and the bottom row features the corresponding refocused images. As we can see in \cref{fig:refocus_res}:\textit{right}, our algorithm does a good job of reconstructing a complex scene when the people are obstructing our view of the majority of the background. The left image in \cref{fig:refocus_res} presents a case of low textured regions (blue curtain and floor) and shows that the algorithm can handle such a situation quite well due to the piece-wise planar prior. In the center indoor scene we can observe the quality of the reconstruction in the presence of significant detail. In the reconstructions there are some portions of the image which are not reconstructed though it is belongs to a dynamic object eg: some portions of upper body in \cref{fig:refocus_res}:\textit{right}. This happens when there is not enough paralallax and rays cannot reach the background and is a function of baseline separation. However, we can recover the majority of the static scene as seen in the images.

\begin{table*}[ht!]
\centering
\medskip
\begin{tabular}{|l|lll|lll|lll|}
\hline
 & \multicolumn{3}{c|}{Less Dynamics (A)} & \multicolumn{3}{c|}{Moderate Dynamic (B)} & \multicolumn{3}{c|}{More Dynamics (C)} \\ \cline{2-10} 
\multicolumn{1}{|c|}{Algorithm} & \begin{tabular}[c]{@{}l@{}}ATE{[}m{]}\\ mean , median\end{tabular} & \begin{tabular}[c]{@{}l@{}}RTE\\ {[}\%{]}\end{tabular} & \begin{tabular}[c]{@{}l@{}}RRE\\ {[}$^{\circ}$/m{]}\end{tabular} & \begin{tabular}[c]{@{}l@{}}ATE{[}m{]}\\ mean , median\end{tabular} & \begin{tabular}[c]{@{}l@{}}RTE\\ {[}\%{]}\end{tabular} & \begin{tabular}[c]{@{}l@{}}RRE\\ {[}$^{\circ}$/m{]}\end{tabular} & \begin{tabular}[c]{@{}l@{}}ATE{[}m{]}\\ mean , median\end{tabular} & \begin{tabular}[c]{@{}l@{}}RTE\\ {[}\%{]}\end{tabular} & \begin{tabular}[c]{@{}l@{}}RRE\\ {[}$^{\circ}$/m{]}\end{tabular} \\ \hline
ORBSLAM2 Mono & \textbf{0.021 , 0.019} & \textbf{3.474} & \textbf{0.404} & 0.399 , 0.019 & 50.987 & 1.238 & 0.987 , 1.000 & 138.24 & 10.798 \\
ORBSLAM2 Stereo & 0.792 , 0.739 & 106.917 & 8.624 & 0.563 , 0.584 & 151.94 & 10.51 & 1.028 , 1.148 & 126.78 & 9.057 \\
DynaSLAM & 0.068 , 0.067 & 5.671 & 0.724 & 0.057 , 0.044 & 6.22 & \textbf{0.494} & 0.065 , 0.046 & 5.126 & 0.846 \\
Light Field front-end & 0.022 , 0.021 & 3.492 & 0.439 & \textbf{0.023 , 0.018} & \textbf{3.66} & 0.57 & \textbf{0.021 , 0.020} & \textbf{3.943} & \textbf{0.733} \\ \hline
\end{tabular}
\caption{Comparison of the ATE [m], RTE(\%) and RRE($^{\circ}$/m) of various visual SLAM algorithms with Light Field front-end in dynamic scenes. The lowest error values are shown in bold font for convenience}
\label{tab:rmse_tab}
\end{table*}
%\textbf{0.019}  , 0.024, 1.000
% 0.739, , 0.584, , 1.148
% 0.067, , 0.044, , 0.024
%, 0.021, \textbf{0.018}, \textbf{0.020}

\begin{figure*}[ht!]
\centering
\begin{tabular}{ccc}
\subfloat[]{\includegraphics[width = 0.29\linewidth,clip, trim={2cm 4.9cm 1cm 3cm}]{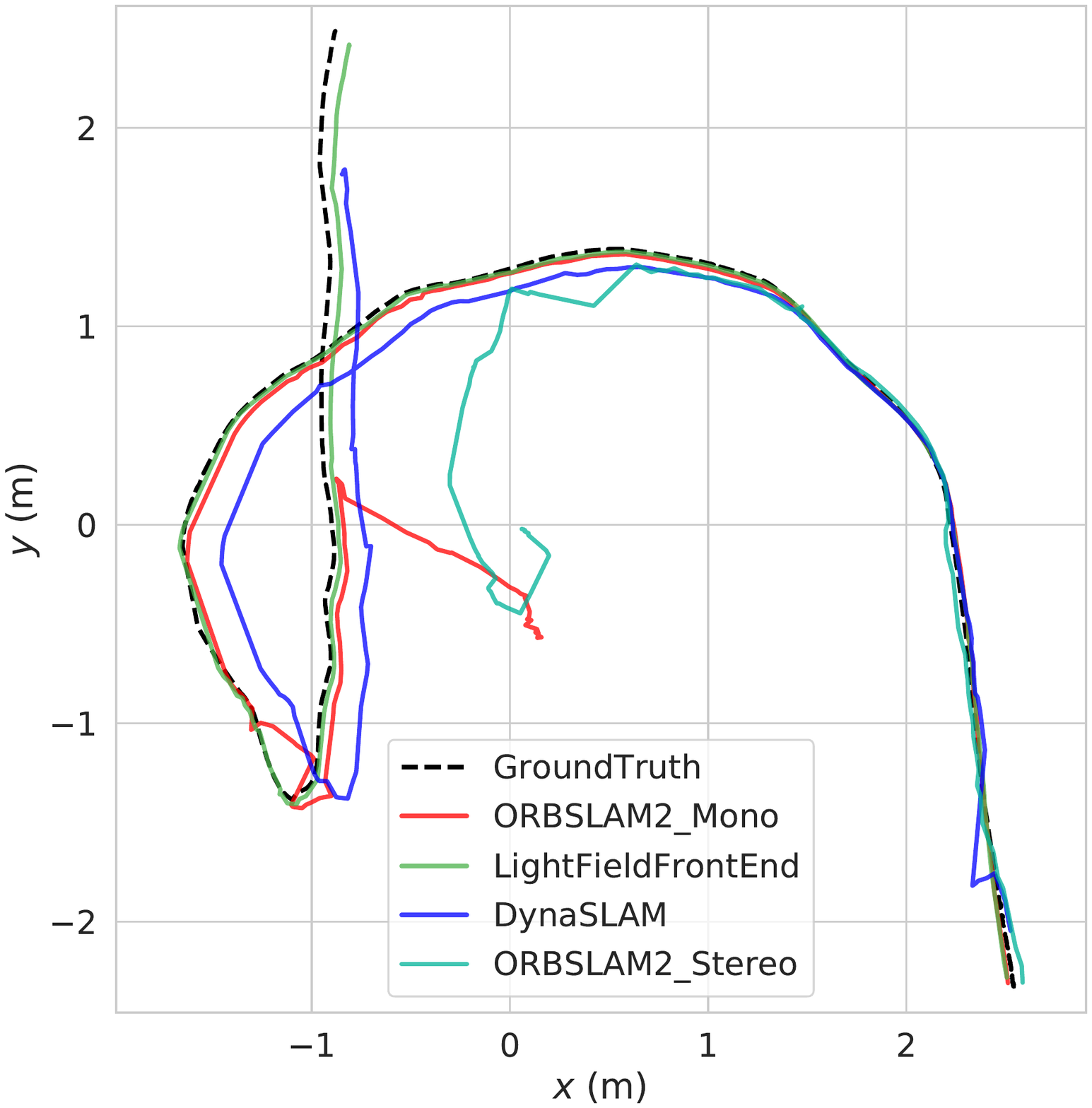}} &
\subfloat[]{\includegraphics[width = 0.3\linewidth, clip, trim={2.1cm 0cm 9.2cm 0cm}]{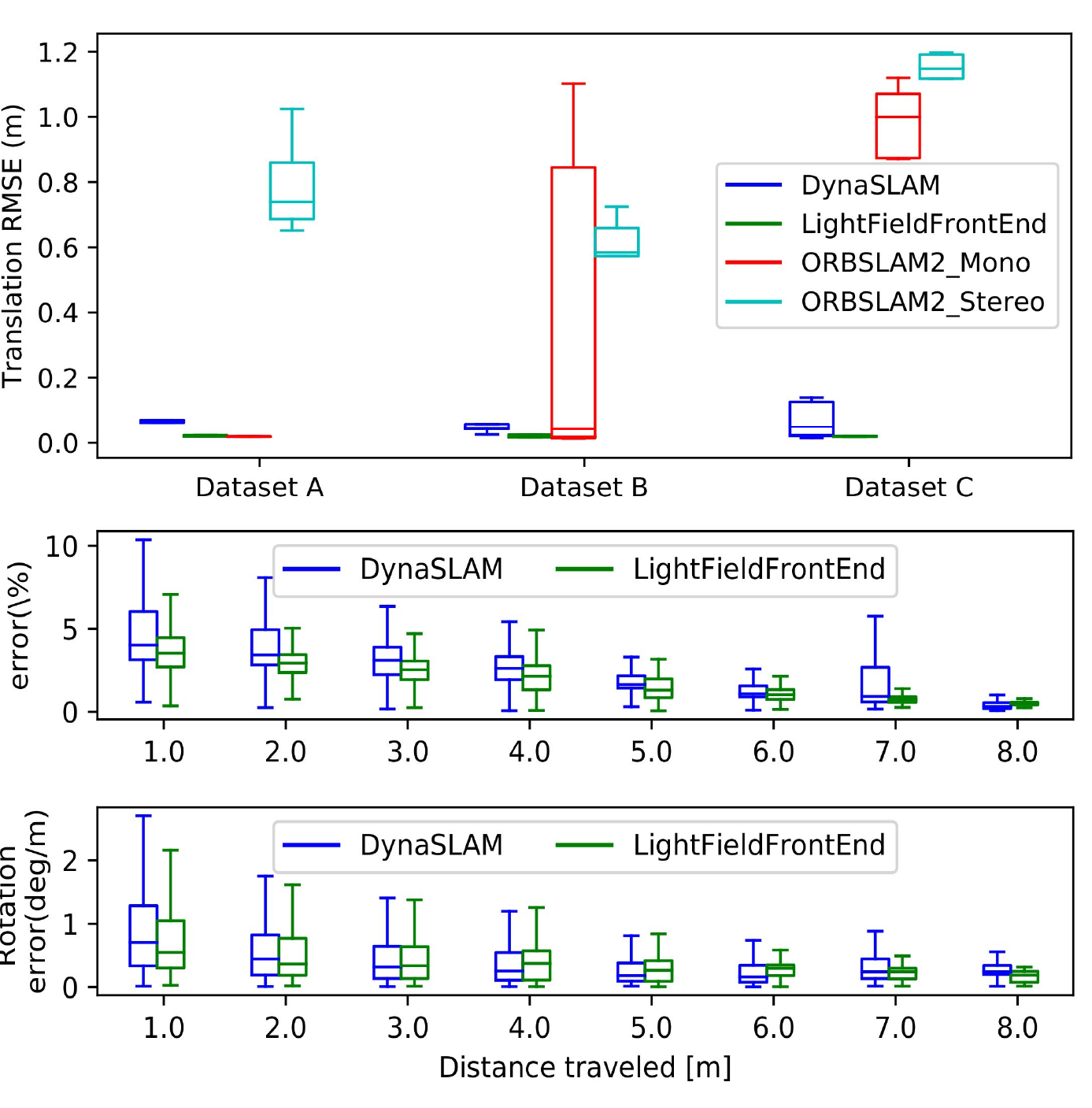}} &
\subfloat[]{\includegraphics[width = 0.29\linewidth,clip, trim={2.2cm 4.9cm 1cm 3cm}]{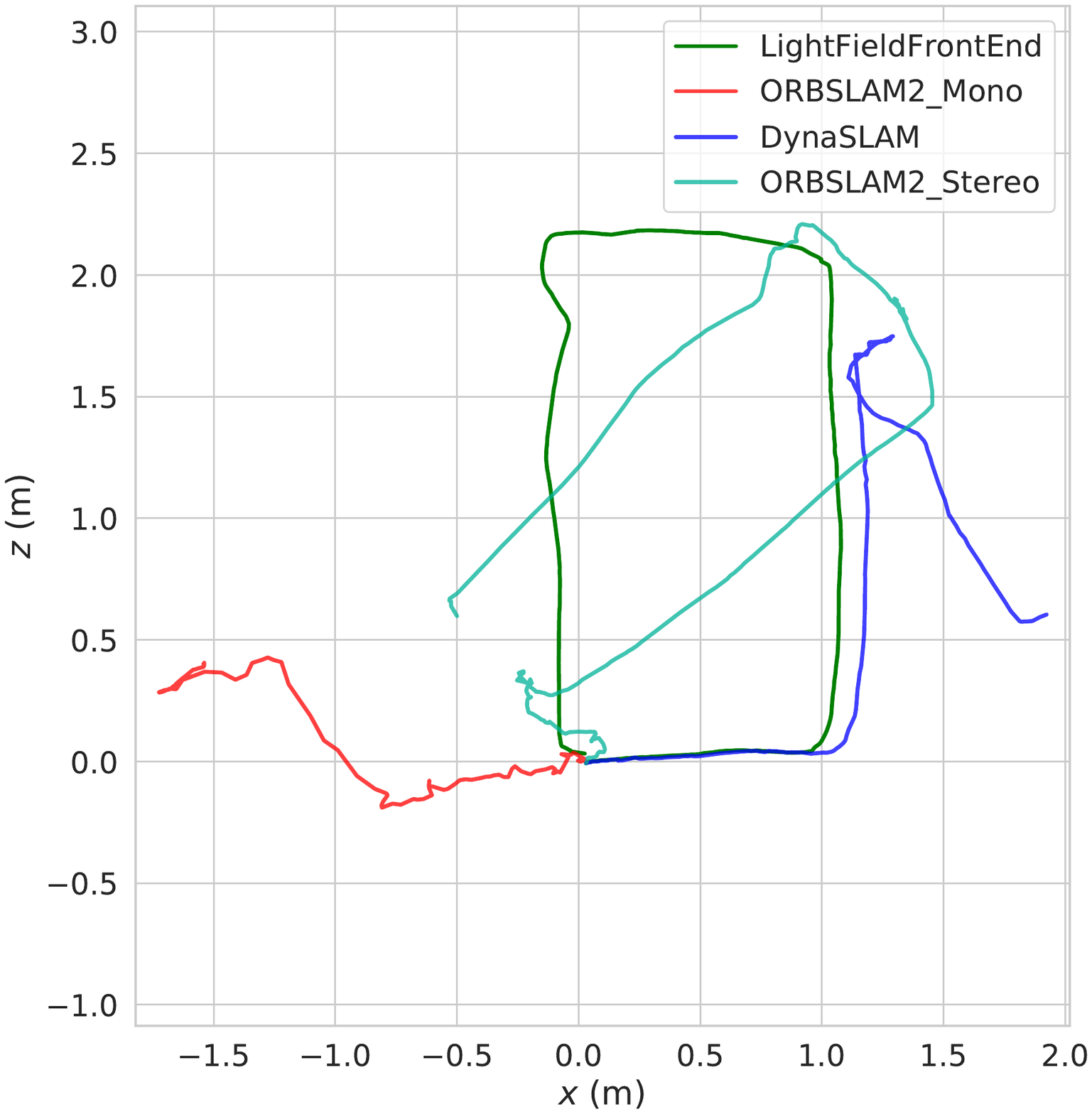}}
\end{tabular}
\caption{(a)Example trajectory plot for ORBSLAM2 Mono, Stereo, DynaSLAM and Light Field front-end compared against the ground truth obtained from optitrack mocap system. (b) \textit{top:} The Absolute Trajectory Error (ATE) for the datasets A,B,C.\textit{center:} The overall RTE (\%) and \textit{bottom:} RRE ($^{\circ}$/m) for different sub trajectory lengths. (c) Trajectories of ORBSLAM2 Mono, Stereo and DynaSLAM compared with that of Light Field front-end for the dataset with loop closure.}
\label{fig:three graphs}
\end{figure*}
\textbf{\textit{Quantitative Analysis}}
%\textcolor{magenta}{Figure 2 : overlay of trajectories with ground truth
%Table1 : Absolute trajectory RMS error 
%Table2: Time analysis 
%Figure 3 : Loop closure failures
%Number of trackable key points??? 
%}
The Light Field front-end is compared with other state of the art SLAM algorithms viz., ORBSLAM2\cite{Mur-Artal2017a} for baseline comparison and DynaSLAM\cite{Bescos2018a} as it also performs static image synthesis in dynamic environments via inpainting. However, the inpainted images are not used for tracking. We demonstrate the effectiveness of our method by comparing the accuracy of pose estimation using Absolute Trajectory Error (ATE) and Relative Pose Error (RPE) metrics. We run the algorithms 5 times on each dataset and take the median errors to account for the non-deterministic nature of system as suggested in \cite{Mur-Artal2017a}and \cite{Bescos2018a}. The evaluation is conducted on three indoor sequences collected with less, moderate and significant amounts of dynamic objects by roughly moving along the same path in each sequence. 

The ATE and RPE for translational and rotational components for the three cases are computed using the RPG trajectory evaluation package\cite{zhang2018tutorial} and are shown in \cref{tab:rmse_tab}. %\cref{fig:three graphs}(a) shows the comparison between ground truth trajectory and the estimated trajectories of various algorithms being compared for one of the datasets.
In \cref{fig:three graphs} (a) the estimated trajectories are aligned and scaled with respect to the ground truth and overlaid on a plot. We can see that Light Field front-end estimates the trajectory closest to the ground truth amongst other algorithms. We observe that ORBSLAM2 Monocular version shows slightly better accuracy than the Light Field front-end in the sequence with less dynamic content due to its robust initialization and outlier rejection strategy. Even though the initialization phase helps with accurate tracking it also often delays bootstrapping in the presence of dynamic objects. This results in missing parts of the estimated trajectory. The performance deteriorates as the proportion of dynamic content increases because the outlier rejection is no longer effective. ORBSLAM2 stereo shows highest errors across all the datasets because the system is initialized from the first frame causing the dynamic keypoints to be tracked.

Both DynaSLAM and Light Field front-end outperform ORBSLAM2(Mono and Stereo) as the dynamic content increases. Our Light Field front-end achieves better accuracy than DynaSLAM in all the datasets except for the relative rotational error (RRE) in dataset b. However we achieve a lower overall RTE and RRE per meter computed for all the datasets when compared to DynaSLAM as shown in \cref{fig:three graphs} (b). While DynaSLAM maintains consistent ATE and RTE values across the three datasets, we can see from \cref{fig:three graphs} (b) that the spread of ATE increases with the dynamic content. This is because in DynaSLAM the features corresponding to the dynamic objects are discarded and only the remaining features are used in the low-cost tracking module for pose estimation. This can introduce errors in tracking when the dynamic features dominate the scene. The Light Field front-end on the other hand demonstrates consistent performance across datasets because instead of discarding dynamic key points associated with segmented pixels we recover the static background, thus gaining the support of static features. 

We further prove that our method is effective in loop closures in the presence of dynamic objects as shown in the \cref{fig:three graphs}(d). This sequence is a rectangular path recorded with people present in the view at the start of the loop. These people, "dyamic objects", move as the trajectory progresses and finally go out of view when we come back to finish the loop. Dynamic objects introduce errors in pose estimation at different points along the trajectory. Monocular ORBSLAM2 experiences extreme errors due to bad initialization. Stereo ORBSLAM2 shows errors initially and DynaSLAM loses tracking at the middle of the trajectory; Both of them recover at a later point but none of the other algorithms achieve loop closure when we come back to the starting point because the feature distribution has changed.%\footnote{DEMO videoINK}.\\
\begin{table}[h]
    \centering
    \begin{tabular}{|c|c|c|c|}
        \hline
         Dataset & depth map [ms] & refocusing[ms] & Total Avg Time[ms]\\
         \hline
         Dataset A & 153.354 & 13.297 & 208.717 \\
         \hline
         Dataset B & 155.482 & 14.357 & 233.918\\
         \hline
         Dataset C & 159.569 & 14.518 & 246.831\\
         \hline
    \end{tabular}
    \caption{Average time taken by various processing steps.}
    \label{tab:time_analysis}
\end{table}
%\begin{table}[h]
%    \centering
%    \begin{tabular}{|c|c|c|c|}
%        \hline
%         dataset & depth map [ms] & refocusing[ms] & Total Avg Time[ms]\\
%         \hline
%         dataset a & 204.44 & 13.7 & 287 \\
%         \hline
%         dataset b & 215.67 & 15.6 & 324.6\\
%         \hline
%         dataset c & 225.28 & 16 & 350\\
%         \hline
%    \end{tabular}
%    \caption{Average time taken by various processing steps.}
%    \label{tab:time_analysis}
%\end{table}
Finally, we report the average computational time for carrying out the different stages of the front-end processing for various sequences in \cref{tab:time_analysis}. To speed up processing we refocus only the dynamic pixels as the other pixels already have the static background intensities. Thus, the computational complexity scales with respect to the amount of dynamic content. Due to the GPU implementation we can achieve speed of 4.35 fps facilitating close to real-time performance.
\section{Conclusion}
\label{sec:conclusions}
We presented a Light Field SLAM front-end where we compute the depth map and synthetic aperture image of the static background using semantic segmentation of apriori dynamic objects. The synthesized image allows us to extract static features occluded by dynamic objects resulting in better tracking accuracy and robustness. This approach performs well even when the scene is dominated by dynamic objects, eliminates the need for an initialization phase, and enables parallel processing resulting in close to real-time executions at ~3 fps on a laptop with a GPU. We show promising results by evaluating our algorithm on real-world datasets collected with dynamic content. In the future, we would like to tightly couple the apriori semantic knowledge with the dynamics of the scene to detect actively moving objects and develop a pose graph back-end for dynamic Light Field SLAM.
\section*{ACKNOWLEDGMENTS}
This work was funded in part by ONR Grant N00014-19-1-2131. We would like to thank S Ramalingam, A Techet and A Bajpayee for their guidance and NU Field Robotics lab for helping with data collection.
%%%%%%%%%%%%%%%%%%%%%%%%%%%%%%%%%%%%%%%%%%%%%%%%%%%%%%%%%%%%%%%%%%%%%%%%%%%%%%%%
\bibliographystyle{IEEEtran}

\bibliography{references}

\end{document}